# Convex Point Estimation using Undirected Bayesian Transfer Hierarchies

**Gal Elidan**
Computer Science Dept.
Stanford University
Stanford, CA 94305
galel@cs.stanford.edu

**Ben Packer**
Computer Science Dept.
Stanford University
Stanford, CA 94305
bpacker@cs.stanford.edu

**Geremy Heitz**
Computer Science Dept.
Stanford University
Stanford, CA 94305
gaheitz@cs.stanford.edu

**Daphne Koller**
Computer Science Dept.
Stanford University
Stanford, CA 94305
koller@cs.stanford.edu

## Abstract

When related learning tasks are naturally arranged in a hierarchy, an appealing approach for coping with scarcity of instances is that of transfer learning using a hierarchical Bayes framework. As fully Bayesian computations can be difficult and computationally demanding, it is often desirable to use posterior point estimates that facilitate (relatively) efficient prediction. However, the hierarchical Bayes framework does not always lend itself naturally to this maximum a-posteriori goal. In this work we propose an undirected reformulation of hierarchical Bayes that relies on priors in the form of similarity measures. We introduce the notion of "degree of transfer" weights on components of these similarity measures, and show how they can be automatically learned within a joint probabilistic framework. Importantly, our reformulation results in a convex objective for many learning problems, thus facilitating optimal posterior point estimation using standard optimization techniques. In addition, we no longer require proper priors, allowing for flexible and straightforward specification of joint distributions over transfer hierarchies. We show that our framework is effective for learning models that are part of transfer hierarchies for two real-life tasks: object shape modeling using Gaussian density estimation and document classification.

## 1 Introduction

Many learning algorithms estimate a parametric model that is intended to generalize well over test datasets. The robustness and success of such methods rely, in large part, on the availability of sufficiently many training instances. In many applications, however, the availability of data is limited and alternative sources of information must be used.

An appealing approach for coping with this scenario involves what is called *transfer learning* (Thrun, 1996; Caruna, 1997), in which data from "similar" tasks/distributions is used to compensate for the sparsity of training data in our primary class or task. When learning a model for the shape of a giraffe, for example, we would like to take advantage of available instances of llamas; when learning to predict the topic of a document, we would like to take advantage of available documents of similar topics. In this paper, we consider the problem of transfer learning in the context of both density estimation and classification.

One general purpose approach for transfer learning is the simple *shrinkage* method designed to improve the estimates of ill-defined problems (e.g., (Carlin and Louis, 1996; McCallum et al., 1998)). This approach involves first learning the parameters of each task/distribution independently and then smoothing the parameters of the most fine-grained tasks (e.g., giraffe, deer) toward the parameters of the coarser-grained tasks (e.g., quadruped).

A more principled approach for transfer learning is the hierarchical Bayes framework, where similar models are related via an existing but unknown prior distribution over the common parameters, and Bayes rule is used to compute posterior parameter estimates (see, for example, Gelman et al. (1995)). This approach is conceptually elegant and effective in many settings, allowing the individual parameters to conform more or less to the prior belief over their values defined by the "parent" task/distribution. For example, it is natural to informally think of the shape distribution of a quadruped as a prior for the shape distribution of a giraffe or a deer. It can also be shown that, asymptotically, using the hierarchical Bayes approach is better than learning related distributions independently, particularly when the prior classes have complex structure (Baxter, 1997).

In practice, as full Bayesian computations can be both difficult and computationally demanding, it is often desirable to perform point estimation of the maximum a-posteriori (MAP) parameters, and use these parameters for (relatively) efficient prediction. Hierarchical Bayes techniques, however, have been designed primarily with Bayesian com-putation in mind and are often not geared toward efficient point estimation. In particular, since many common priors such as the Dirichlet or normal-inverse-Wishart are not convex with respect to their arguments, the MAP objective does not lend itself naturally to efficient and optimal optimization. In addition, due to mathematical technicalities, the parametrization of such conjugate priors is often problematic for high-dimensional domains (see Section 2).

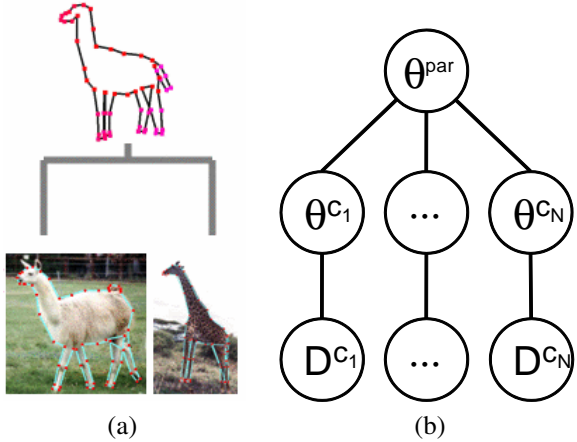

Figure 1: (a) Illustration of a simple transfer hierarchy where the distributions over the shapes of a pair of mammals share a common prior distribution (e.g., long-necked mammal). (b) An example of a hierarchical Bayes parametrization. The joint distribution over the data (observations of the leaf classes) and parameters has the form of $P(\Theta_{par})\prod_i P(\Theta^{C_i} \mid \Theta_{par})P(D^{C_i} \mid \Theta^{C_i})$.

In this work we aim to preserve the probabilistic appeal of the hierarchical Bayes setting while applying it to the problem of posterior point estimation in high-dimensional transfer hierarchies. As in the standard hierarchical Bayes approach, we define a single probabilistic objective that combines fitting the data with terms that encourage soft similarity between several related distributions and a common parent distribution (see Figure 1). Unlike the standard hierarchical Bayes approach, we replace the prior distributions with positive similarity measures and write our joint probability distribution as an undirected Markov random field (MRF). This can be viewed as defining an improper prior over the parameters of all classes. This model falls under the broad definition of probabilistic abstraction hierarchies (PAHs) of Segal et al. (2001), but refines this general framework by defining a coherent joint distribution over all parameters in the hierarchy.

The hierarchical Bayes perspective also motivates an important extension of the joint distribution that is not captured by a standard shrinkage approach (e.g., (McCallum et al., 1998)) or by the PAH model (Segal et al., 2001). We introduce the notion of "degrees of transfer" along the edges of the hierarchy, allowing for different weights on different components of the similarity measures. These weights afford the possibility of more intelligent sharing in that similarity need not be equally strong throughout the hierarchy. In Section 4, we show how this idea is made concrete and how the weights can be automatically learned using the same optimization-based approach.

The reformulation of a hierarchical Bayes objective as an undirected model with improper priors has several benefits. First, specifying the joint distribution is straightforward, as it only requires a choice of a divergence measure between the parameters of the parent and child distributions. Second, multi-tier hierarchies present no additional complications, as conjugacy is not required for the goal of point estimation. Most importantly, if we choose a convex divergence term (e.g., L1, L2, $e$-insensitive, Kullback-Leibler), the problem of finding posterior point estimates is convex for many learning scenarios and can be solved using standard gradient based techniques.

We make our approach concrete for two very different settings. First, we consider a continuous distribution setting in which the shapes of different mammals share a common quadruped shape prior. Second, we construct a multi-tier hierarchy of naive Bayes models for document classification into topics based on the newsgroup dataset. In both cases we demonstrate that our high-dimensional undirected hierarchical Bayes model, joined with convex point estimation of posterior parameters, is superior to distributions learned independently as well as to the shrinkage approach.

## 2 Generative Hierarchical Bayes

Our goal is to define a framework for flexible transfer learning in a class hierarchy. In this section, we briefly describe the standard hierarchical Bayes approach for this scenario. In the next section, we present our alternative formulation.

We begin with a given hierarchy over a set of related learning tasks/classes $\mathcal{C}$ (for example, Figure 1(b)). At the lowest level are the leaf classes $c \in \mathcal{L}$ for which we observe instances $\mathcal{D}^c$. For convenience, we assume that $\mathcal{D}^c$ only exists for the nodes at the leaves of the hierarchy. [1] For each class $c$ (except the root), $par(c)$ denotes its parent class.

The hierarchical Bayesian framework views this hierarchy as a directed probabilistic model in which the class parameters $\Theta^c$ are generated according to the distribution $P(\Theta^c \mid \Theta^{par(c)})$ and the data is generated according to $P(\mathcal{D}^c \mid \Theta^c)$ for $c \in \mathcal{L}$. This induces a joint distribution over the observed data and all class parameters as follows:

$$P(\mathcal{D}, \Theta) = \prod_{c \in \mathcal{L}} P(\mathcal{D}^c \mid \Theta^c) \times \prod_{c \in \mathcal{C}} P(\Theta^c \mid \Theta^{par(c)}) \quad (1)$$

where for the root class $P(\Theta^c \mid \Theta^{par(c)}) \equiv P(\Theta^c)$.

---
[1] Our general formulation in Section 3 can easily incorporate observations of any class in the hierarchy.

Once this directed model is defined, we are typically interested in the posterior distribution of the parameters $P(\Theta \mid \mathcal{D})$, or the marginal of this posterior for a subset of the parameters. In the *empirical Bayes* approach, tailored for the common case of a two-level hierarchy, a point estimate for the root parameters is found first, and then full posteriors $P(\Theta^c \mid \mathcal{D}, \Theta^{root})$ are computed for the leaf distributions (see Gelman et al. (1995) for more details).

As noted in Section 1, it is often preferable to compute point estimates of the MAP parameters rather than use full posteriors for the practical purpose of prediction. That is, we want to compute

$$\Theta^* = \underset{\Theta}{\text{argmin}} -\log P(\mathcal{D}, \Theta),$$

where

$$\log P(\mathcal{D}, \Theta) = \sum_{c \in \mathcal{L}} \log P(\mathcal{D}^c \mid \Theta^c) + \sum_{c \in \mathcal{C}} \log P(\Theta^c \mid \Theta^{par(c)}) \quad (2)$$

By design, many standard hierarchical Bayes approaches lend themselves toward the fully Bayesian treatment, and therefore make limiting choices. In particular, priors are generally chosen to be in conjugate families to allow for the computation of posteriors in closed form. However, many common conjugate priors such as the Dirichlet or normal-inverse-Wishart are not convex with respect to the parameters, thus not allowing for efficient point estimation.

In addition, the requirement that the priors be valid distributions that integrate to 1 can lead to untenable restrictions. For example, the conjugate prior for the multivariate Gaussian is the normal-inverse-Wishart distribution, which has a degrees of freedom parameter corresponding to the strength (pseudocounts) of the prior. In order to ensure that it is a proper distribution, the pseudocounts must be at least as large as the dimension $d$ of the data. Thus, in the case of high-dimensional data with few instances, this requirement forces the strength of the prior to overwhelm the signal in the data. For example, in Section 5 we consider 120-dimensional data representing the shape of various mammals, with up to only 15 instances per class. Even if we perform dimensionality reduction so that the data is only 20-dimensional, using a normal-inverse-Wishart in this case would mean that the weakest allowable prior would be at least as strong as the data. Indeed, most hierarchical Bayes applications are limited to the scenario in which the higher level priors are low-dimensional and relatively simple (e.g., (Gelman et al., 1995; Blei et al., 2003)), rather than a rich class distributions in their own right.

## 3 Undirected Transfer Hierarchies

As mentioned above, many of the difficulties associated with the standard hierarchical Bayes approach result from the design of the framework for full posterior estimation. Thus, for the purpose of point posterior estimation, we propose an alternative formulation that is more amenable to the use of standard optimization techniques. We start by describing the general approach and then provide two different scenarios for which it is made concrete. In Section 4 we present an extension to the basic approach that allows for greater flexibility in the way that parameters are shared.

### 3.1 Basic Framework

We begin our reformulation by writing Eq. (2) as an objective that has a more general form (and that can be made equivalent to Eq. (2) given the appropriate instantiation):

$$F_{\text{joint}}(\Theta; \mathcal{D}) = -\sum_{c \in \mathcal{L}} \mathcal{F}_{\text{data}}(\mathcal{D}^c, \Theta^c) + \beta \sum_{c \in \mathcal{C}} \text{Div}(\Theta^c, \Theta^{par(c)}) \quad (3)$$

Here, $\mathcal{F}_{\text{data}}(\mathcal{D}^c, \Theta^c)$ is a data-dependent objective (e.g., the probability of $\mathcal{D}^c$ given $\Theta^c$) that encourages the parameters of each class to represent the data well. $\text{Div}(\Theta^c, \Theta^{par(c)})$ is a divergence or dissimilarity function over the child and parent parameters that encourages the parameters of linked classes to be similar. The weight $\beta$ determines the overall trade-off between these competing objectives.

In the analysis that follows, we consider divergence functions that decompose independently over each parameter in $\Theta^c$. This includes many dissimilarity metrics, including all norms. Using this assumption, we write:

$$F_{\text{joint}}(\Theta; \mathcal{D}) = -\sum_{c \in \mathcal{L}} \mathcal{F}_{\text{data}}(\mathcal{D}^c, \Theta^c) + \beta \sum_{c \in \mathcal{C}} \sum_{i} \text{Div}(\theta_i^c, \theta_i^{par(c)}) \quad (4)$$

Note that we can more generally take $\Theta_i^c$ to be (possibly overlapping) subsets of $\Theta^c$. For the sake of simplicity, however, we only consider the fully decomposed form.

There are several points to note about our objective in Eq. (4). First, it has the general form of the log-probability of a *Markov random field* (MRF) (Pearl, 1988), and if its integral is finite, then $\frac{1}{Z} \exp\{-F_{\text{joint}}(\Theta; \mathcal{D})\}$ defines a coherent joint probability function over the data and parameters of all classes in the hierarchy. Thus, this formulation is in fact a hierarchical Bayes setting where we make use of unnormalized priors in the form of $\exp\{-\text{Div}(\theta_i^c, \theta_i^{par(c)})\}$. Second, the penalty term is a function of both the child and parent parameters. This means that the penalty can be reduced by either parameter moving towards the other. It is this feature that differentiates us from empirical Bayes.

While Eq. (3) and Eq. (4) may appear as deceptively naive rewritings of Eq. (2), this view actually has important practical ramifications. The first benefit is the ease with which relatedness may be defined. Natural choices include the L2 distance (corresponding to a local unnormalized Gaussian prior), the L1 distance (corresponding to a local unnormalized Laplacian prior), or a (smoothed) $\epsilon$-insensitive loss (which gives a constant penalty for all values below $\epsilon$).

The second benefit is that we can now use inference/optimization techniques that have been developed for energy objectives that take on the form of Eq. (3). Our goal is to learn a point estimate for the parameters that is the mode of the posterior distribution. That is, given $\mathcal{D}^c$ for all classes, we want to learn the parameters $\Theta^*$ such that

$$\Theta^* = \operatorname*{argmin}_\theta F_{\text{joint}}(\Theta; \mathcal{D})$$

Appealingly, for *any* convex divergence function $\text{Div}(\Theta^c, \Theta^{par(c)})$, this entire objective is convex in $\Theta$, when the data objective $\mathcal{F}_{\text{data}}(\mathcal{D}^c, \Theta^c)$ (e.g., likelihood) is concave, as is typically the case for many learning scenarios. This allows us to find the mode of optimal parameters using straightforward gradient ascent techniques; in the experiments below, we use the Polak-Ribiere conjugate gradient algorithm (Boyd and Vandenberghe, 2004). Note that our approach scales well with large amounts of data, as the sample size affects only the one-time collection of sufficient statistics for use in $\mathcal{F}_{\text{data}}$.

Below, we describe how the reformulation of the hierarchical Bayes model is made concrete for two different transfer learning scenarios. In Section 5 we present experiments using these instantiations for two real-life scenarios.

### 3.2 Gaussian Density Estimation

Suppose we wish to model the data of class $c$ using a multivariate Gaussian distribution parametrized by mean $\mu^c$ and covariance $\Sigma^c$. The data objective for this class is naturally given by the (regularized) log-likelihood:

$$\begin{aligned}\mathcal{F}_{\text{data}}(\mathcal{D}^c, \Theta^c) &\equiv \ell(\mathcal{D}^c; \Theta^c) \quad (5)\\ &= \sum_m^{M_c} \log \mathcal{N}(\mathbf{x}[m] \mid \mu^c, \Sigma^c + \alpha \mathcal{I}),\end{aligned}$$

where $\mathbf{x}[m]$ is the $m$th instance in $\mathcal{D}^c$ (which is of size $M_c$), $\mathcal{I}$ is the identity matrix, and $\alpha$ is the standard regularizing "ridge" term used to avoid issues of singular matrices and overfitting in the regime of small amounts of data.

This objective is not concave in the covariance parameter $\Sigma^c$, but becomes concave if the Gaussian is parametrized using the *inverse* covariance matrix, also known as the precision matrix. In this case, the log-likelihood is:

$$\begin{aligned}\ell(\mathcal{D}^c; \Theta^c) = &-\frac{1}{2}\sum_m^{M_c}(\mathbf{x}[m] - \mu)^T K(\mathbf{x}[m] - \mu)\\ &-\frac{M}{2}\log \det K + C \quad (6)\end{aligned}$$

where $K^{-1} = \Sigma^c + \alpha \mathcal{I}$.

Unfortunately, the Gaussian log-likelihood is *not* jointly concave in the mean vector and precision matrix. However, it *is* concave in each part independently. As a result, we can iteratively optimize our objective in two phases, one for each set of parameters. This process generally converges to a local optimum in a few iterations.

For the divergence function, we use the L2 norm over the mean and diagonal precision parameters, while we do not tie the off-diagonal terms to the parent parameters.

### 3.3 Discrete Bayesian Networks

In contrast to the case of continuous Gaussian distributions, we consider the setting in which each class model is a Bayesian network over discrete-valued data. Note that in this scenario, each node in the hierarchy is a Bayesian network, and we assume that the networks have the same structure and therefore the same parametrization $\Theta^c$.

We consider the specific case of table CPDs (conditional probability distributions) using a multinomial distribution. To avoid positivity and normalization constraints, we use a log-space representation of the multinomial distribution

$$P(x \mid pa_X; \theta^c) = \frac{\exp(\theta^c_{x, pa_X})}{\sum_{x'}\exp(\theta^c_{x', pa_X})},$$

where $x$ and $pa_X$ are specific values for the variable $X$ and its parents. Note that the notion of the parents of $X$ is in the context of the Bayesian network that sits inside each node in the hierarchy; this is distinct from the notion of parents of the nodes themselves in the hierarchy.

We now define the data objective to be the log-likelihood of the data of a class $c$ given the parameters $\Theta^c$ of the corresponding Bayesian network:

$$\begin{aligned}\mathcal{F}_{\text{data}}(\mathcal{D}^c, \Theta^c) &\equiv \ell(\mathcal{D}^c; \Theta^c)\\ &= \sum_m^{M_c}\sum_{X \in \mathcal{X}} \log P(x[m] \mid pa_X[m]; \theta^c)\\ &= \sum_{X \in \mathcal{X}}\sum_x (N_c\{x, pa_X\} + \alpha) \cdot\\ &\quad \{\theta^c_{x, pa_X} - \log \sum_{x'}\exp(\theta^c_{x', pa_X})\} \quad (7)\end{aligned}$$

where the sufficient statistic $N_c\{x, pa_X\}$ is the number of times the values $(x, pa_X)$ occur in the dataset $\mathcal{D}^c$, and $\alpha$ is the corresponding pseudo-count of the Dirichlet prior distribution used to regularize the likelihood. For the divergence function, we use the L2 norm over all parameters $\theta^c$.

## 4 Degree of Transfer Coefficients

Our model thus far can be viewed as a probabilistic reinterpretation of the PAH framework of Segal et al. (2001).

Having motivated our model as an undirected reformulation of the hierarchical Bayes joint distribution, we can use this novel perspective to suggest additional modifications that are natural to this setting. In particular, we can refine the single weight $\beta$ that corresponds to prior strength in Eq. (4) and introduce distinct $\lambda_i^{c,par(c)}$ terms at each edge $(c, par(c))$ in the hierarchy. These *degree of transfer* (DOT) coefficients represent how much we want each of the child parameters to be near its corresponding parent parameter. Note that, as with the $\text{Div}(\Theta^c, \Theta^{par(c)})$ functions, we can define DOT coefficients for groups of parameters rather than individual ones.

To make the benefit of different degrees of transfer coefficients concrete, consider a document classification task as a motivating example. Suppose that the child class is the 'biology textbook' topic with a small number of training instances and the parent class is the 'science textbook' topic, with a significantly larger number of training instances. The frequency of the word 'experiment', represented by the multinomial parameter $\theta_{\text{experiment}}$, is likely to be similar for both the child and parent class. On the other hand, the parameter $\theta_{\text{gene}}$ corresponding to the word 'gene' is likely to be misrepresented at the parent class. We therefore stand to gain from penalizing the distance between child and parent more for $\theta_{\text{experiment}}$ and less for $\theta_{\text{gene}}$.

Incorporating the degree of transfer coefficients $\lambda_i^{c,par(c)}$, our objective now becomes:

$$F_{\text{joint}}(\Theta, \Lambda; \mathcal{D}) = -\sum_{c \in \mathcal{L}} \mathcal{F}_{\text{data}}(\mathcal{D}^c, \Theta^c)$$
$$+ \beta \sum_{c \in \mathcal{C}} \sum_i \frac{1}{\lambda_i^{c,par(c)}} \text{Div}(\theta_i^c, \theta_i^{par(c)}) \quad (8)$$

This allows us the flexibility to turn sharing on and off in a continuous fashion for the various edges in the hierarchy. For instance, if we set $\lambda_i^{c,par(c)}$ near zero, we are effectively forcing the parameters to agree, whereas if we make it large, we are allowing as much flexibility as is necessitated by the likelihood term.

Naively, we might try treating these factors as parameters of the joint transfer objective of Eq. (8) and optimize with respect to both the class parameters and transfer factors. Upon examination of the objective in Eq. (8), however, we notice that the optimal value occurs when all transfer parameters approach infinity ($\lambda_i^{c,par(c)} \to \infty$) and all model parameters are set to their independent maximum likelihood values. This should not come as a surprise, as the transfer parameters play the role of a prior strength, which is usually estimated using cross validation or some other external means. Obviously, while such an approach can be used to estimate the global weight $\beta$, we cannot hope to cross validate the large number of $\lambda$ parameters. How then can we estimate the parameter values in a meaningful way?

Our first option is to choose a reasonable *constant* value for $\lambda_i^{c,par(c)}$ before we optimize the parameters $\Theta$. We describe an empirical Bayes bootstrap approach for doing so in Section 4.1. In Section 4.2, we describe a different approach that places a prior over the $\lambda$ coefficients and optimizes them along with the model parameters $\Theta$.

## 4.1 Undirected Empirical Bayes Estimation

In order to choose an appropriate constant value for each $\lambda_i^{c,par(c)}$, recall that it represents the inverse strength of the divergence penalty between a child and parent parameter. Thus, it should be assigned a lower value if we expect the parameter to be near its parent, and a higher value if we expect it to be far. One way to quantify this is to consider randomly sampled subsets of the child data $\mathcal{D}^c$: if the maximum likelihood estimate for a parameter $\theta_i^c$ is consistently close to the corresponding parent parameter $\theta_i^{par(c)}$ across the sampled datasets, then we want to encourage higher similarity; otherwise, we set a lower penalty coefficient.

Formally, we define a random variable $\delta_{i,c} = \theta_i^c - \theta_i^{par(c)}$ and wish to estimate the expected variance of this variable *across all possible datasets*. We use the bootstrap approach of Efron and Tibshirani (1993) to approximate this expectation. That is, for each class, we create $K$ random datasets of size equal to the original training set by uniformly sampling with replacement from the original dataset. For each of these sets we then estimate the standard (regularized) maximum likelihood parameters $\theta^c$ and $\theta^{par(c)}$, and compute the empirical variance of the difference $\delta_{i,c}$ across the $K$ trials, which we denote by $\hat{\sigma}^2_{c,par(c),i}$. We then use this disparity measure to set the transfer coefficients:

$$\lambda_i^{c,par(c)} = \hat{\sigma}^2_{c,par(c),i}$$

Using this approach together with, for example, the quadratic penalty, our objective becomes

$$F_{\text{joint}}(\Theta; \mathcal{D}, \Lambda) = -\sum_{c \in \mathcal{L}} \mathcal{F}_{\text{data}}(\mathcal{D}^c, \Theta^c)$$
$$+ \beta \sum_{c \in \mathcal{C}} \sum_i \frac{(\theta_i^c - \theta_i^{par(c)})^2}{\hat{\sigma}^2_{c,par(c),i}}. \quad (9)$$

Under this formulation, we see that the penalty terms in the objective represent a product of Gaussian priors over the difference between the parent and child values of parameter $\theta_i$. If we were to fix the parent parameter to its bootstrap value, this would reduce to empirical Bayes using the parent bootstrap estimates as parameters of the Gaussian prior. We therefore call this approach the undirected empirical Bayes estimation of the transfer factors.

The approach proposed in this section is appealing from a computational perspective for two reasons. First, the complexity of the estimation of the transfer factors is proportional to the complexity of the estimation of the ML parameters of each class independently. Second, given the transfer factors that are computed only once, the objective Eq. (9) is convex for many typical learning tasks and choices of the divergence function $\text{Div}(\Theta^c, \Theta^{par(c)})$ and can thus be efficiently optimized.

## 4.2 Hyperprior-Based Estimation

A second approach is to estimate each DOT transfer coefficient along with the model parameters themselves. To ensure that these DOT coefficients are not driven to infinity, we can add a prior that pushes the factors towards smaller values. Specifically, we can add an inverse-Gamma prior to the transfer factors, forcing them to be positive. This is a natural choice since the inverse-Gamma is the conjugate prior distribution for Gaussian variances, and we can (loosely) interpret $\lambda$ as the variance of a Gaussian prior for the model parameter. In this case, our objective expands to:

$$F_{\text{joint}}(\Theta, \Lambda; \mathcal{D}) = \sum_c -\ell(\mathcal{D}^c; \Theta^c)$$
$$+ \beta \sum_{c \in \mathcal{L}} \sum_i \frac{(\theta_i^c - \theta_i^{par(c)})^2}{\lambda_i^{c,par(c)}}$$
$$- \sum_{c \in \mathcal{C}} \sum_i \log G^{-1}(\lambda_i^{c,par(c)}) \quad (10)$$

where $G^{-1}$ is the density function of the inverse-Gamma distribution. We choose the parameters of the inverse-Gamma distribution such that the mean is equal to the bootstrapped value $\hat{\sigma}^2_{c,par(c),i}$ as in Section 4.1.

Importantly, if our data objective (e.g., log-likelihood) function is concave, our entire objective is jointly convex in the model parameters ($\theta$'s) and the DOT coefficients ($\lambda$'s). This is an attractive aspect of our formulation, and allows for efficient and guaranteed optimal inference of the MAP values for the parameters.

## 5 Experimental Evaluation

In this section, we demonstrate how our reformulation of the hierarchical Bayes model as a Markov random field allows us to effectively transfer across classes. We consider the task of density estimation for multivariate Gaussian shape models as well as a document classification task. In all experiments we compare our hierarchical approach to a **CV Reg** model for which each class is learned independently. The parameters of **CV Reg** are chosen to maximize the regularized likelihood, and the regularization coefficients are determined using cross-validation. To provide an additional baseline, we also compare to a **Shrinkage** approach (see below for the details in each application).

### 5.1 Mammal Shape Model Learning

We adopt the notion of object shape modeled by a Gaussian distribution as in Elidan et al. (2006) and attempt to learn a shape model for several classes of mammals. Each instance is a set of 60 landmarks from hand-outlined images represented as a 120-dimensional vector (using the x and y coordinates of each landmark). Our dataset consists of 40 such instances of Elephants, Bison, Rhinos, Giraffes, and Llamas. Since these classes are well-represented on a lower-dimensional manifold, we use PCA to reduce the dimensionality to 20. The objective formulation and covariance matrix regularization are explained in Section 3.2.

We evaluate the benefit of transferring between pairs of mammal classes. We train each method using $N = \{3, 5, 10, 15\}$ instances for each child class, evaluating the log-likelihood of 20 test instances for each child class. We report averages using five fold cross-validation.

To demonstrate the usefulness of the degree of transfer (DOT) coefficients, we consider three variants of our method. **Bootstrap** uses bootstrapping to estimate the DOT coefficients as described in Section 4.1. **Hyperprior** uses the approach described in Section 4.2 to define a hyperprior distribution over the DOT coefficients and estimate the DOT coefficients together with the other model parameters. In both these cases we set the global weight $\beta$ to 1. To evaluate whether including the DOT coefficients provides any advantage, the **CV Const** method does not use DOT coefficients, instead cross-validating $\beta$ in the range of $10^{-6}$ to 1 (a typical value range of the coefficient when using **Bootstrap**). To make it as competitive as possible, we cross-validate specifically for each mammal pair and for each number of training instances $N$.

We compare these variants of our method to the independent **CV Reg** baseline, which uses cross-validated regularization as described in Section 3.2. Our experiments showed that a standard **Shrinkage** approach performs very poorly in this case due to the fact that linearly interpolating covariance matrices yields nonsensical (and possibly invalid) distributions. We attempted a version of shrinkage over the diagonal entries of the covariance alone, but this also resulted in poor performance, as the appropriate interaction between the variance (diagonal) and covariance (off-diagonal) elements is disturbed. We therefore omit the **Shrinkage** variant from the graphs reported below.

Figure 2 shows the difference in test log-likelihood between our methods and the **CV Reg** baseline. (a) shows the benefit of our **Hyperprior** approach for all mammal pairs. The benefit of our approach is evident and typically lies in the range of $5-10$ bits per instance on test data. Out of 10 mammal pairs, our method is not superior to **CV Reg** in only a single case and only when the number of training instances of each class is 3.

In Figure 2(b) we consider the rhino-bison mammal pair, which is "close" in shape, and compare the success of the different variants of our method. The benefit of having independent DOT coefficients is clear as **CV Const** is in-ferior, due to the fact that samples are scarce and cross-validation is not robust. Figure 2(c) shows an example of the giraffe-rhino mammal pair that are farther apart. For this pair, **CV Const** was at least 100 bits per instance worse than **CV Reg** for all training set sizes, so it is omitted from the graph. Although the **Hyperprior** method is somewhat more advantageous than the **Bootstrap** approach, we note that overall the methods are quite competitive.

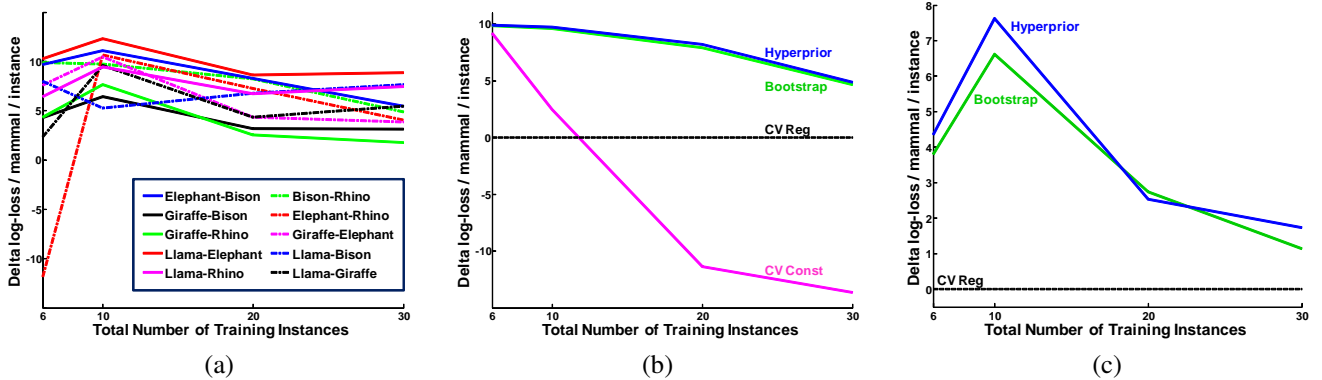

Figure 2: Results for transfer between mammal pairs. The numbers reported are the average across 5 randomized folds of the improvement in test log-likelihood per mammal per instance (y-axis) over the regularized **CV Reg** baseline, as a function of number of samples (x-axis). (a) shows average improvement in performance for all mammal pairs of our method with a **Hyperprior** over the degree of transfer (DOT) coefficients; (b) compares **Hyperprior** with our **Bootstrap** estimation of the DOT coefficient as well as the cross-validated **CV Const** for the "similar" bison and rhino pair; (c) is the same as (b) for the "dissimilar" llama and rhino pair. In (c), the **CV Const** method is omitted, as it is at least 100 bits per instance worse than **CV Reg** for all training set sizes. Typically, **Hyperprior** training time was approximately 90 seconds.

### 5.2 Newsgroup Posting Classification

In order to demonstrate our method on a larger and more involved hierarchy than the simple mammal pairs, we use the Newsgroup dataset, which consists of 18,827 newsgroup postings drawn from 20 distinct newsgroups. For the experiments described here, we use a hierarchy over 15 classes, gathered into the abstract classes Religion, Politics, Vehicles, Sports, and Computers, as in McCallum et al. (1998). This creates a hierarchy with 15 leaf nodes, five middle level node, and one root node, as shown in Figure 3(a).

Documents are tokenized and all tokens occurring only one time are removed to produce a corpus with 55,989 unique words. When individual experiments are performed, we use 100 test documents, and all words not present in the training or test set for that particular experiment are removed (to improve efficiently and clarity of the results), typically leaving around 20000 words. We report average results using 5 fold cross validation.

In addition to the regularized **CV Reg** model, we implemented a slightly simplified variant of hierarchical **Shrinkage** (McCallum et al., 1998) for which, starting from the root of the node, each node's parameters are shrunk toward its parent using a single parameter learned by k-fold cross validation. Due to the size of the optimization problem in this case, we use a simplified **Undirected HB** version of our approach that does not make use of the individual DOT coefficients and where we set the global weight $\beta$ to 1.

For both the baselines as well as our method, we use a naive Bayes model as an instantiation of the approach described in Section 3.3. A document $d$ in this framework is modeled as a bag of words, where $d_i$ indicates the number of times that word $i$ appears in the document. We model each class as a probability distribution over words, where each word in the document is considered independently:

$$P(d \mid c) = \prod_i P(w_i \mid c)^{d_i}.$$

We use a multinomial distribution as in Section 3.3 for $P(w_i \mid c)$. The full data objective for this model is:

$$\mathcal{F}_{\text{data}}(\mathcal{D}^c, \Theta^c) \equiv \sum_{d \in \mathcal{D}^c} \sum_i (d_i + \alpha) \cdot \{\theta_i^c - \log \sum_j \exp(\theta_j^c)\}$$

We evaluate each method using its classification rate on test instances. To highlight the importance of regularization (through the parameter $\alpha$), we also present the baseline **Likelihood** that simply uses the maximum likelihood parameters for each class without regularizing. Figure 3 shows the consistent advantage of our **Undirected HB** approach over all baselines. The advantage of the regularized **CV Reg** over **Likelihood** is also clear, demonstrating the extent to which even simple regularization, once cross-validated, can be beneficial. This also explains why the regularized **CV Reg** model achieves a better classification rate than the **Shrinkage** method that uses Laplacian smoothing (setting $\alpha = 1$), as in McCallum et al. (1998).

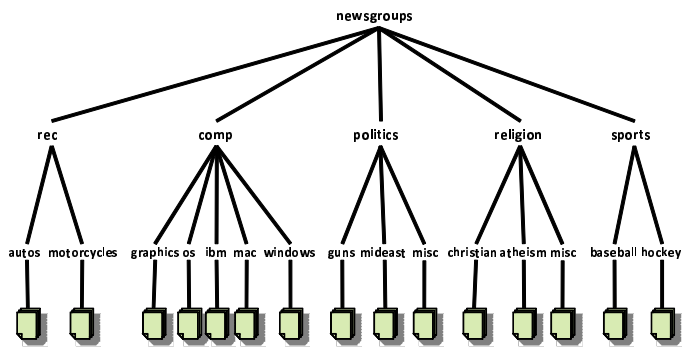
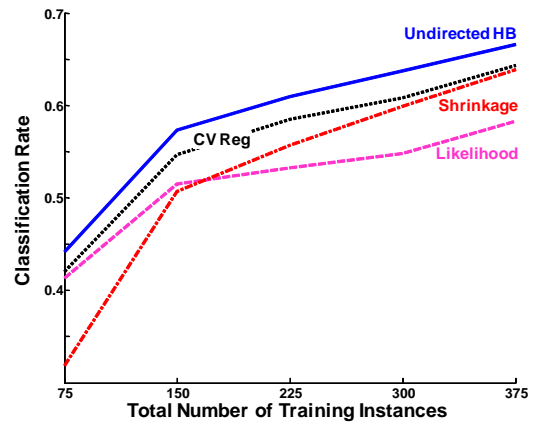

Figure 3: (left) A hierarchy of topics for the Newsgroup dataset adopted from McCallum et al. (1998). (right) Results for the 15 class Newsgroup hierarchy. Shown is average classification accuracy across 5 random train and test splits (y-axis) as a function of number of samples (x-axis). Our **Undirected HB** approach is compared to the recursive **Shrinkage** approach, the regularized **CV Reg** model, and to non-regularized **Likelihood**. The typical **Hyperprior** training time was approximately 30 minutes, comparable to that of **Shrinkage**.

## 6 Summary and Future Directions

In this work, we proposed an undirected reformulation of the hierarchical Bayes framework that replaces proper priors with similarity measures for the purpose of posterior point estimation in high-dimensional transfer hierarchies. We introduced the notion of "degree of transfer" weights that afford our model flexibility not available in other standard hierarchical methods, and suggested two approaches for automatically estimating these weights. Finally, we showed how the general approach can be applied to the scenario of shape modeling using a multivariate Gaussian density and to document topic classification. In both settings, we demonstrated the superiority of our approach over both independent learning and a shrinkage-based competitor.

The benefits of our reformulation are threefold. First, while retaining the hierarchical Bayes appeal of using a prior as the mechanism for transfer, our method is straightforward and allows for straightforward specification of the objective. Second, by writing a Markov random field objective, we can use convex optimization techniques to find point estimates of the posterior distributions for a large range of learning scenarios and similarity measures. Third, our framework allows for different degrees of transfer for different parameters so that some parts of the distribution can be transferred to a greater extent than others.

There are several extensions to this work to explore. The first is learning the structure of transfer itself, at the class and parameter level. The PAH framework (Segal et al., 2001) allows the structure of the hierarchy to be learned; one could adapt their approach to both identify hierarchy edges and discover which parameters (e.g., those that correspond to a coherent part of a mammal) transfer together. Furthermore, unlike the classical hierarchical Bayes approach, there is nothing in our framework that prevents a class from having multiple parent classes. Such hierarchies are common in many settings (including the Wordnet hierarchy and the GO hierarchy in biology), and it would be interesting to explore whether this added flexibility can be of benefit in a Bayesian learning task.